\documentclass{IOS-Book-Article}
\usepackage{mathptmx}
\usepackage{soul}\setuldepth{article}
\usepackage{booktabs}
\usepackage{multirow}
\usepackage{graphicx}
\usepackage{caption}
\usepackage{subcaption}
\usepackage{amsmath}

\def\hb{\hbox to 11.5 cm{}}

\begin{document}

\def\thepage{}

\begin{frontmatter}              % The preamble begins here.

%\pretitle{Pretitle}
%\title{Drafting Ontologies Using FrODO - The Frame-based Ontology Design Outlet}
\title{Automatically Drafting Ontologies from Competency Questions with FrODO}

%\markboth{}{May 2022\hb}
%\subtitle{FrODO - Frame-based Ontology Desing Outlet}

\author[A,B]{\fnms{Aldo} \snm{Gangemi}},
\author[A]{\fnms{Anna Sofia} \snm{Lippolis}},
\author[A]{\fnms{Giorgia} \snm{Lodi}},
\author[A]{\fnms{Andrea Giovanni} \snm{Nuzzolese}\thanks{Corresponding Author: Andrea Giovanni Nuzzolese; E-mail:
andreagiovanni.nuzzolese@cnr.it}\thanks{The authors are sorted alphabetically as they equally contributed to this paper.}}

\runningauthor{Aldo Gangemi et al.}
\runningtitle{Automatically Drafting Ontologies from
Competency Questions with FrODO}
\address[A]{Institute of Cognitive Sciences and Technologies, Italian National Research Council (ISTC-CNR), Via San Martino della Battaglia 44, 00184, Rome, Italy}
\address[B]{Department of Classical and Italian Philology, University of Bologna, Via Zamboni 32, 40126, Bologna, Italy}
%\address[B]{Short Affiliation of Second Author and Third Author}

\begin{abstract}
We present the Frame-based ontology Design Outlet (FrODO), a novel method and tool for drafting ontologies from competency questions automatically. Competency questions are expressed as natural language and are a common solution for representing requirements in a number of agile ontology engineering methodologies, such as the eXtreme Design (XD) or SAMOD. FrODO builds on top of FRED. In fact, it leverages the frame semantics for drawing domain-relevant boundaries around the RDF produced by FRED from a competency question, thus drafting domain ontologies. We carried out a user-based study for assessing FrODO in supporting engineers for ontology design tasks. The study shows that FrODO is effective in this and the resulting ontology drafts are qualitative.

%evaluation of the tool is introduced in the paper with a discussion of the main obtained results. 

%Starting from competency questions derived from user stories defined for modelling an application domain, the tool relies on the machine reading method FRED in order to construct a knowledge graph. This can be successively navigated in order to produce a preliminary version of an OWL ontology, responding to the competency questions. The resulting OWL ontology can be then loaded in ontology editors such as Protégé for further manipulations.
%A user evaluation of the tool is introduced in the paper with a discussion of the main obtained results. 
\end{abstract}

\end{frontmatter}
%\markboth{June 2021\hb}{June 2021\hb}
%\thispagestyle{empty}
%\pagestyle{empty}

\section{Introduction}
\label{sec:intro}
Competency questions~\cite{Gruninger1995} (CQs) expressed as natural language are a common solution for representing requirements in a number of agile ontology engineering methodologies, such as the eXtreme Design~\cite{Presutti2009} (XD) or SAMOD~\cite{SAMOD2016}. In such methodologies most effort lies in the design of ontology modules able to address the CQs that have been previously identified, which is a fully manual activity. Hence, it represents a clear bottleneck for agile methodologies. This is fairly evident in situations in which ontology drafts need to be shared among ontology engineers and domain experts or stakeholders for incremental refinements and/or knowledge exchange. Accordingly, it is utmost important that those ontology drafts, although tentative, must comply with best design practices, e.g. Ontology Design Patterns, and properties, e.g. cognitive ergonomics, transparency and flexibility.
In this paper we present the Frame-based Ontology Design Outlet (FrODO), which is a novel method and Web tool for automatically drafting OWL ontologies from CQs. FrODO builds on and benefits from FRED~\cite{Gangemi2017} for {\em machine reading}~\cite{Etzioni2007} aimed at gathering RDF from natural language. FRED is a formal machine reader that produces RDF graphs from text, which are (i) domain- and task-independent, and (ii) designed according to the frame semantics~\cite{Fillmore2006} and ontology design patterns~\cite{Gangemi2009}. Hence, FrODO extends FRED specifically on the case of CQs by tailoring the RDF produced by FRED into domain ontologies by leveraging its formal representation based on the frame semantics. The domain ontologies produced by FrODO are drafts that can be used to feed agile ontology design methodologies. Accordingly, in this work we investigate the following research questions:
\begin{itemize}
    \item {\em RQ1}: Is frame semantics fair to be exploited for generating well structured ontology drafts from CQs?
    \item {\em RQ2}: Do ontology engineers benefit from ontology drafts that are automatically produced from CQs in their ontology engineering tasks?
\end{itemize}

In order to address the aforementioned research questions we carried out a user-based study. The participants to the study were asked to use FrODO during the design starting from a set of given CQs. The quality of the resulting ontologies was measured with well known structural metrics. The effectiveness of FrODO was measured in terms of usability within the context of a an ontology engineering workflow.
The rest of the paper is organised as it follows: Section~\ref{sec:relwork} surveys some related works; Section~\ref{sec:frodo} describes the methodology implemented by FrODO; Section~\ref{sec:usereval} describes the evaluation, presents and discusses the results; finally, Section~\ref{sec:concl} presents our conclusions and future works.

\section{Related Work}
\label{sec:relwork}
Competency questions (CQs) are questions in natural language that define and constrain the scope of knowledge represented in an ontology. As such, they are used in ontology engineering as requirements useful to evaluate an ontology based on its ability to answer each question, particularly in several agile methodologies, such as, the TOVE enterprise modeling approach~\cite{GrunFox1995}, eXtreme Design (XD) in \cite{Presutti2009}, SAMOD~\cite{SAMOD2016}, On-To-Knowledge~\cite{Sure2004}). All the aforementioned methodologies can be defined as test driven~\cite{Keet2016,Wizniewski2019}. Namely, they assess the commitment of ontologies to the requirements by converting CQs into queries, e.g. SPARQL, DL queries, etc. For example, \cite{Carriero2019} is a Web application designed for providing the XD methodology with a testing toolbox based on the representation of CQs to SPARQL queries. In this context~\cite{Wisniewski2018} provides a solution for generating SPARQL queries from CQs. Our solution is meant to be used as a component of the aforementioned agile methodologies. However, it does not automatise the generation of queries for testing purposes, but provides a solution for drafting an ontology from its associated CQs automatically. The latter point can be seen an Ontology Learning and Population (OL\&P) task~\cite{Cimiano2006,Al2020}. Examples of such methods include~\cite{Cimiano2005,Witte2010,Tanev2006}. Most of these solutions are implemented on top of machine learning methods. Hence, they are typically data hungry, i.e. they require large corpora, sometimes manually annotated, in order to learn rules for ontology automatic construction. Such rules are defined through a training phase that can take a long time. On the contrary, our solution does not depend on any training and is unsupervised. Other approaches to OL\&P use either lexico-syntactic patterns~\cite{Hearst1992}, or hybrid lexical-logical techniques~\cite{Voelker2008}. However, to the best of our knowledge no practical tools have emerged so far for doing it automatically while preserving high quality of results. Finally~\cite{Gangemi2017} is a formal machine reader able to transform natural language text into domain-independent formal structured knowledge represented as RDF/OWL. Our solution build on top of FRED for generating domain-dependent ontologies from CQs.

\section{The Frame-based ontology Design Outlet}
\label{sec:frodo}
\begin{figure}[!hbt]
\centering
\includegraphics[width=1\textwidth]{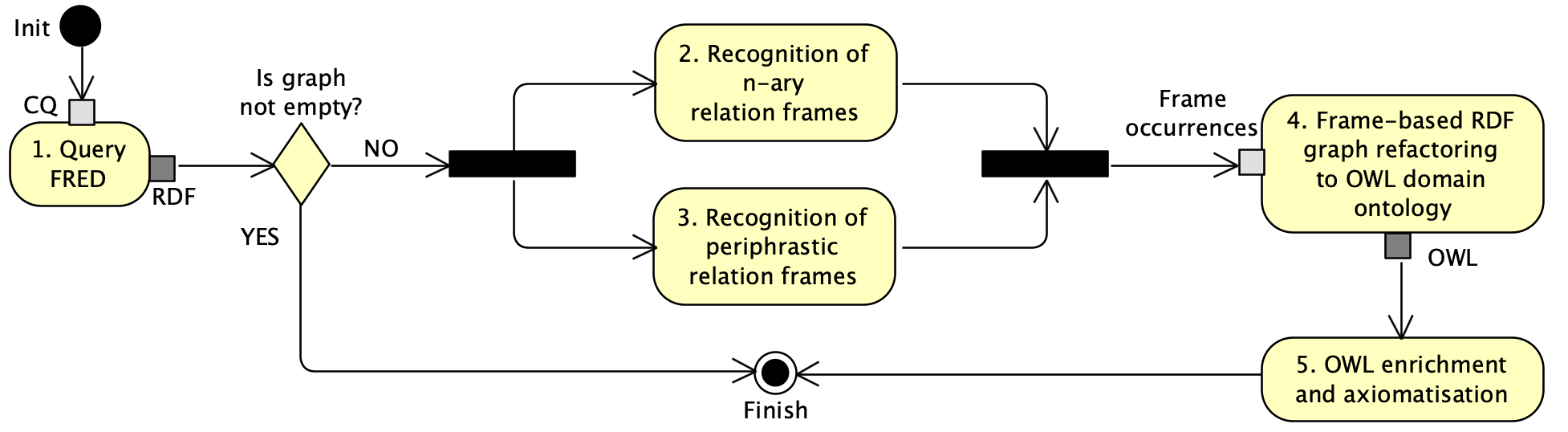}
\caption{Methodology implemented by FrODO.}
\label{fig:methodology}
\end{figure}

We generate ontologies from CQs by refactoring the RDF graph produced by FRED. This is done by means of graph traversal strategies that exploit the frame semantics~\cite{Fillmore2006}. We assume that frames and frame arguments, as represented by FRED, are the key tools to leverage on for drawing domain-relevant boundaries around the classes and properties produced by FRED. This enables the generation of domain ontology drafts. In fact, on one hand frames convey general meaning, i.e. they are bound in FRED to VerbNet frames that are broader and domain-independent concepts. On the other hand, frame arguments enable a solution for specialising such concepts in a specific domain with peculiar knowledge gathered from text directly. Furthermore, frame roles (e.g. agent, patient, theme, etc), that link a frame to its arguments, can be used for introducing peculiar naming conventions, annotations, and axioms in the draft ontologies that are fine-grained to a domain. Frames are represented by FRED in two alternative ways, that is, (i) as $n$-ary relations and (ii) periphrastic relations.
Figure~\ref{fig:methodology} depicts the methodology implemented by FrODO through the UML notation for activity diagrams. In such a figure it is fairly evident how FrODO extends FRED (cf. activity 1) by adding frame recognition (cf. activities 2 and 3) and domain ontology generation and enrichment (cf. activities 4 and 5). The activities if Figure~\ref{fig:methodology} are detailed in the subsequent sections.

\subsection{Drafting ontologies from $n$-ary relations}
\begin{figure}[!t]
\centering
\includegraphics[width=0.85\textwidth]{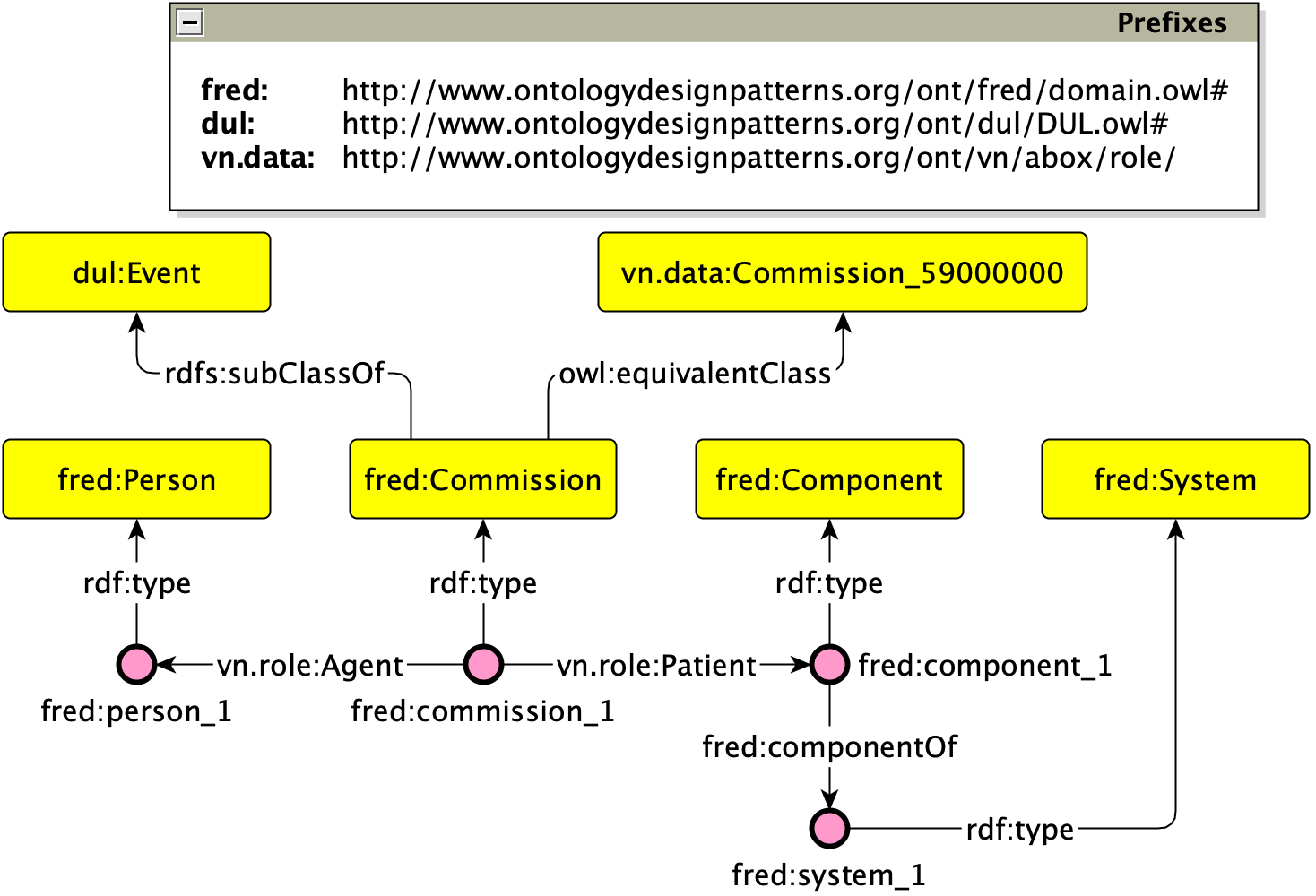}
\caption{RDF graph produced by FRED for the text \textit{``Who commissioned a component of a system?''}. The graph is drawn with the Graffoo notation~\cite{Falco2014}}
\label{fig:fred_graph}
\end{figure}

In the case of $n$-ary relations, a frame occurrence is represented by an individual of a class, which is, in turn, a sub-class of \texttt{dul:Event}. The sub-classing of \texttt{dul:Event} might be not direct. Hence, we need to traverse the \texttt{rdfs:subClassOf} axioms transitively. This can be exemplified with the RDF graph produced by FRED for the CQ \textit{``Who commissioned a component of a system?''}, which is depicted in Figure~\ref{fig:fred_graph}. In such a Figure a frame occurrence based on the $n$-ary pattern is identified by the individual \texttt{fred:commission\_1}. This individual is a valid frame occurrence as Equation~\ref{eq:trav1} is matched. In fact, \texttt{fred:commission\_1} is an instance of \texttt{fred:Commission}, which is, in turn, a sub-class of \texttt{dul:Event}. The RDF graph depicted in Figure~\ref{fig:fred_graph} and the frame identified are possible outputs for the activities 1 and 2 in Figure~\ref{fig:methodology}, respectively.

%Accordingly, the first step of the refactoring strategy of FrODO is focused on recognising frame occurrences. Equation~\ref{eq:trav1} represents the set $F_{n-ary}$ whose members are all frame occurrences. A frame occurrence in FRED is represented by an individual of a class, which, in turn, is a sub-class of dul:Event. The sub-classing of dul:Event might be not direct. Hence, we need to traverse the rdfs:subClassOf axioms transitively.

\vspace{-0.5pt}
\begin{equation}
\label{eq:trav1}
F_{\mbox{{\em n-ary}}} \equiv \{f|\mbox{ } f \in C \mbox{ }\wedge\mbox{ } C \in \mbox{\texttt{owl:Class}} \mbox{ }\wedge\mbox{ } C \subseteq^{+} \mbox{\texttt{dul:Event}}\}
\end{equation}

The arguments of an $n$-ary relation are utmost important for providing the context to an identified frame. In fact, the arguments are the individuals participating in an event, i.e. the frame represented as an $n$-ary relation. These arguments are linked to the event by means of binary predicates, i.e. the frame roles. According to the frame semantics based on VerbNet implemented by FRED possible roles are: 
\begin{itemize}
\item agentive roles, such as \texttt{vn.role:Agent}, \texttt{vn.role:Actor}, etc.; 
\item passive roles, such as \texttt{vn.role:Patient}, \texttt{vn.role:Patient1}, etc.;
\item thematic roles, such as \texttt{vn.role:Theme}, \texttt{vn.role:Theme1}, \texttt{vn.role:Experiencer};
\item oblique roles, such as \texttt{vn.role:Asset}, \texttt{vn.role:Time}, etc.;
\item roles based on periphrastic relations, such as \texttt{fred:at}, \texttt{fred;composeOf}, \texttt{fred:personOf}, etc..
\end{itemize}

Domain classes in a draft ontology are defined by constructing compound terms from frame occurrences represented with the $n$-ary pattern and their arguments. In this construction the arguments with a passive role are core. In fact, FrODO retrieves the name of the class typing the argument of an $n$-ary that plays a passive role. This class name is then concatenated with the name of the class of the frame occurrence (i.e. the target $n$-ary) itself. The latter name is first declined into gerund form. FrODO does not take into account agentive roles for this construction as they are typically bound to individuals that are typed with classes that are too broader or not constrained to a specific domain, e.g. \texttt{fred:Agent}, \texttt{fred:Person}, etc. This is due to the fact that the arguments with agentive roles are typically produced from pronouns, e.g. \textit{``Who''}, \textit{``What''}, \textit{``Which''}, etc., by following a shared pattern for defining CQs~\cite{Ren2014}. On the contrary, arguments with a passive role are typically associated with domain-relevant terms in CQs, e.g. {\em ``component''}. As an example, in Figure~\ref{fig:fred_graph} the type of the frame occurrence is \texttt{fred:Commission}, while the type of its argument playing a passive role is \texttt{fred:Component}. Hence, the new class generated by FrODO for representing a domain-relevant $n$-ary in the draft ontology is \texttt{:ComponentCommissioning}\footnote{In this paper the prefix \texttt{:} is reserved for referring to the namespace used by FrODO.}. The second step of the approach is to define the classes and properties to use as the arguments of the new $n$-ary classes introduced in a draft. This is done by reusing the classes typing the arguments of a frame occurrence in FRED regardless to the specific role played in the $n$-ary relation. Based on our running example, this means that both \texttt{:Component} and \texttt{:Person} are defined as classes in the ontology drafted by FrODO. Then, those classes are linked to the $n$-ary they belong to by defining new object properties in place of the frame roles provided by FRED. These object properties are constructed by applying a naming convention based on the template \texttt{:involves\{ClassName\}}, where \texttt{\{ClassName\}} is replaced by the actual name of the class being an argument of the target $n$-ary with a camel case notation. For example, the object properties \texttt{:involvesComponent} and \texttt{:involvesPerson} are generated for the classes \texttt{:Component} and \texttt{:Person}, respectively. We opt for the term {\em ``involves''} as it evokes the involvement of a participant, i.e. the argument, into a situation, i.e. the frame. The generated object properties are then provided with domain and range axioms. The range axioms are set to the corresponding argument of the $n$-ary, e.g. \texttt{:Component} and \texttt{:Person} are the range of \texttt{:involvesComponent} and \texttt{:involvesPerson}, respectively. On the contrary, domain axioms are set to \texttt{owl:Thing}, hence they are kept open. The association with the corresponding $n$-ary relation is materialised by means of existential restriction axioms defined locally to the class representing the $n$-ary relation following the pattern defined in Equation~\ref{eq:restr}. This Equation allows FrODO to cope with the activity 2 of Figure~\ref{fig:methodology}.

\vspace{-0.5pt}
\begin{align}
\label{eq:restr}
C_{\mbox{{\em n-ary}}} & \sqsubseteq \exists r_{arg}.C_{arg}, & \mbox{{\em n-ary}} \in F_{\mbox{{\em n-ary}}}, \forall \langle p, arg\rangle \mbox{ s.t. } p(\mbox{{\em n-ary}}, arg) \in G_{FRED}
\end{align}

In Equation~\ref{eq:restr} (i) $C_{\mbox{{\em n-ary}}}$ is the class generated by FrODO from the $n$-ary frame occurrence produced by FRED, which belongs to $F_{\mbox{{\em n-ary}}}$; (ii) $C_{arg}$ is the class generated by FrODO for an argument of the $n$-ary in FRED; (iii) $r_{arg}$ is the object property generated by FrODO for an argument of the $n$-ary in FRED; (iv) $p$ is a frame role, i.e. an RDF predicate, that links the $n$-ary frame occurrence and the corresponding argument in the graph produced FRED, which is $G_{FRED}$.
For each generated object property its corresponding inverse is materialised. The naming convention for inverse object properties is based on the template \texttt{:is\{ClassName\}InvolvedIn}. In this template \texttt{\{ClassName\}} is substituted with the actual name of the class being an argument of the target $n$-ary with a camel case notation. For example, \texttt{:isComponentInvolvedIn} and \texttt{:isPersonInvolvedIn} are defined as inverse object properties of \texttt{:involvesComponent} and \texttt{:involvesPerson}, respectively. 
Addiontally, for each generated class a taxonomy is inferred following the compositional semantics. For example, \texttt{:ComponentCommissioning} is declared to be \texttt{rdfs:subClassOf} \texttt{:Commissioning}. Finally, each class and property defined in the draft ontology is associated with a human readable label, i.e. \texttt{rdfs:label}. The following code is the OWL produced by FrODO for the CQ used so far in our running example. The OWL is serialised with the Manchester syntax. All the solutions explained for generating axioms and labels are part of the activity 5 in Figure~\ref{fig:methodology}.

\begin{scriptsize}
\begin{verbatim}
ObjectProperty: involvesComponent
     Annotations: rdfs:label "involves component"
     Domain: owl:Thing
     Range: Component
     InverseOf: isComponentInvolvedIn
ObjectProperty: involvesPerson
    Annotations: rdfs:label "involves person"
    Domain: owl:Thing
    Range: Person
    InverseOf: isPersonInvolvedIn
ObjectProperty: isComponentInvolvedIn
    Annotations rdfs:label "is component involved in"
    Domain: Component
    Range: owl:Thing
    InverseOf: involvesComponent
ObjectProperty: isPersonInvolvedIn
    Annotations: rdfs:label "is person involved in"
    Domain: Person
    Range: owl:Thing
    InverseOf: involvesPerson
Class: Commissioning
    Annotations: rdfs:label "Commissioning"@en
Class: Component
    Annotations: rdfs:label "Component"
Class: ComponentCommissioning
    Annotations: rdfs:label "Component commissioning"@en
    SubClassOf: Commissioning,
                involvesComponent some Component,
                involvesPerson some Person
Class: Person
    Annotations: rdfs:label "Person"
\end{verbatim}
\end{scriptsize}

\subsection{Drafting ontologies from periphrastic relations}
Periphrastic relations occur in FRED when it generates meaningful object properties by concatenating nouns with their corresponding prepositions, e.g. of, with, for, in, etc. For example, in Figure~\ref{fig:fred_graph} a periphrastic relation is \texttt{fred:componentOf}. In fact, it is the result of the concatenation of the noun {\em ``component''} and the preposition {\em ``of''} as they appear in the input text of the CQ. Periphrastic relations are relevant in our scenario as they express frames as binary relations and not in terms of $n$-relations. They can be easily identified in the graph as they are OWL object properties defined by FRED using a local namespace. Equation~\ref{eq:trav2} formalises the set of frames evoked by periphrastic relations as $F_{periphrastic}$, thus providing us a method to address the activity 3 in Figure~\ref{fig:methodology}. The local namespace\footnote{The local namespace can be customised by a user in FRED either by means of its API or Web interface. We remark that FRED uses a number of namespaces and prefixes associated with external ontologies and vocabularies used for representing a text as RDF, e.g. DOLCE, VerbNet, DBpedia, etc. We refer the interested readers to~\cite{Gangemi2017} for more details about FRED.} is used by FRED for producing URIs from the input textual elements. In Figure~\ref{fig:fred_graph} the local namespace is associated with the predix \texttt{fred:}.
\vspace{-0.5pt}
\begin{equation}
\label{eq:trav2}
F_{periphrastic} \equiv \{f|\mbox{ } f \in \mbox{\texttt{owl:ObjectProperty}} \mbox{ }\wedge\mbox{ } namespace(R) = \mbox{\texttt{fred:}}\}
\end{equation}

Hence, we explain how FrODO copes with activities 4 and 5 depicted in Figure~\ref{fig:methodology} for periphrastic relation. In the case of a periphrastic relation the frame is the binary relation itself, which produces an object property, while the arguments are the subject and object of the relation that play the agentive and passive roles, respectively. The object property is generated with a naming convention that follows the template \texttt{:\{relation\}\{ObjectClassName\}} with a camel case notation. In this tamplate (i) \texttt{\{relation\}} is substituted by the name of the periphrastic relation produced by FRED, e.g. \texttt{componentOf}, and (ii) \texttt{ObjectClassName} is substituted by the name of the class typing the individual that plays the passive role in the relation, e.g. \texttt{System}. The object property \texttt{:componentOfSystem} is produced for the periphrastic relation in our running example. The naming convention is based on the assumption that arguments that play a passive role convey domain-peculiar knowledge. Then, two classes are generated from the subject and object of the periphrastic relation. The naming convention is based on the template \texttt{:\{ClassName\}}. In such a template \texttt{\{ClassName\}} is a variable to substitute with the actual value being the name of the class typing either the subject or the object of the periphrastic relation. For example, the classes \texttt{:Component} and \texttt{:System} are produced for the case represented by our example. Accordingly, the object properties resulting from periphrastic relations are enriched with domain and range axioms by following the same rationale adopted for the $n$-ary relation case. That is, for a given object property the domain and range are set to \texttt{owl:Thing} and the class produced from the individual being the object of a periphrastic relation, respectively. In our example, the range of the object property \texttt{:componentOfSystem} is \texttt{:System}. Furthermore, an inverse relation is materialised for each object property produced. The naming convention used for the inverse object properties is based on the template \texttt{:is\{ObjectProperty\}of} with a camel case notation. In such a template \texttt{\{ObjectProperty\}} is a variable which is substituted with the name of the object property that the current one is an inverse of. For example, is \texttt{:isComponentOfSystemOf} is the inverse produced for the object property \texttt{:componentOfSystem}.
Then, the class generated from subject of the periphrastic relation is axiomatised with an existential restriction by applying the pattern defined in Equation~\ref{eq:restr2}.

\vspace{-0.5pt}
\begin{align}
\label{eq:restr2}
S' & \sqsubseteq \exists r_{p}.O', & \mbox{{\em p}} \in F_{\mbox{{\em periphrastic}}}, s \in S, o \in O, p(s,o) \in G_{FRED}
\end{align}

In Equation~\ref{eq:restr2} (i) p is a predicate that recognised as a periphrastic relation and holding between a subject $s$ and an object $o$ is a graph $G_{FRED}$, e.g. the triple $\langle\mbox{\texttt{fred:component\_1}}, \mbox{\texttt{fred:componentOf}},\mbox{\texttt{fred:system\_1}}\rangle$; (ii) $S$ is the class used as the type for the subject $s$, e.g. \texttt{fred:Component}; (iii) $O$ is the class used as the type for the object $o$, e.g. \texttt{fred:System}; $S'$ and $O'$ are the classes generated by FrODO for $S$ and $O$, e.g. \texttt{:Component} and \texttt{:System}; (iv) $r_{p}$ is the binary property produced by FrODO from the input periphrastic relation, e.g. \texttt{:isComponentOfSystemOf}. 

Finally, FrODO adds \texttt{rdfs:label} annotations to generated classes and properties. The following is the OWL produced by FrODO from the periphrastic relation occurring in our running example.

\begin{scriptsize}
\begin{verbatim}
ObjectProperty: componentOfSystem
    Annotations: rdfs:label "component of system"
    Domain: owl:Thing
    Range: System
    InverseOf: isComponentOfSystemOf
ObjectProperty: isComponentOfSystemOf
    Annotations: rdfs:label "is component of system of"
    Domain: System
    Range: owl:Thing
    InverseOf: componentOfSystem
Class: Component
    Annotations: rdfs:label "Component"
    SubClassOf: componentOfSystem some System
Class: System
    Annotations:  rdfs:label "System"@en
\end{verbatim}
\end{scriptsize}

%\subsection{Implementation details}
FrODO is implemented as a Python Web application. Its source code is available on a GitHub repository\footnote{\url{https://github.com/anuzzolese/frodo}} and a running instance is available online\footnote{\url{https://w3id.org/stlab/frodo} that redirects to \url{http://semantics.istc.cnr.it/frodo}}.

\section{Evaluation}
\label{sec:usereval}
\subsection{Experimental setup}

We designed our experiment as a user-based study in order to address {\em RQ1} and {\em RQ2}. Namely, we asked a number of participants to perform an ontology design task starting from a set of identified CQs used as ontological requirements. The ontology design task was composed of two conditions. The first condition, i.e. {\em C1}, aimed at designing an ontology able to answer the provided CQs without the support of FrODO. On the contrary, the second condition, i.e. {\em C2}, aimed at designing an ontology able to answer the provided CQs with the support of FrODO. In both conditions the target ontology editor was Prot\'{e}g\'{e}. In case of the condition C2 the participants were asked to copy a CQ from the list of CQs they were provided with and paste it into the Web interface of FrODO. Then, they were asked to (i) execute FrODO in order to get the resulting ontology and (ii) import such an ontology into Prot\'{e}g\'{e}. This operation had to be executed once per each CQ the participants were provided with. No strategy on the order of CQs to solve for both condition C1 and C2 was imposed to the participants, i.e. they could start from any of the provided CQs by following their preferred order. Additionally, in case of condition C2 (with FrODO), the participants could opt for the implementation strategy the felt more confident with. This means, for instance, a participant might first import all the ontology modules returned by FrODO for each CQs and then refine those modules in the final ontology as a whole. Similarly, a participant might import and refine the ontology modules produced by FrODO for each CQs one-by-one incrementally.
For the selection of the CQs we first identified three common CQ patterns out of the set of 12 CQ patterns observed in literature by~\cite{Ren2014}. The selected CQ patterns capture actors, relations, quantities, and modalities, as well as temporal and spatial elements. Namely the three CQ patterns are: 
\begin{itemize}
\item {\em ``When is [object] [relation]?''}, i.e. pattern P1; 
\item {\em ``What is [object] [relation] [object]?''}, i.e. pattern P2; 
\item {\em ``Who [relation] [object]?''}, i.e. pattern P3. 
\end{itemize}
For each of the three patterns we randomly picked two CQs from the corpus of CQs we defined in the context of the WHOW project\footnote{The Water Health Open knoWledge (WHOW) is a project co-financed by the Connecting European Facilities of  the European Union under grant agreement 2019-EU-IA-0089. Project website: \url{https://whowproject.eu/}}. This allowed us to get two sets of CQs, i.e. {\em S1} and {\em S2} having three CQs each. Table~\ref{tab:cqs} reports the CQs selected along with their corresponding set, identifier, and pattern.
\begin{table}[!t]
\centering
\caption{CQs used as sample in our experiment.}\label{tab:cqs}
\begin{tabular}{lclr}
\toprule                      
\textbf{Set} & \textbf{ID} & 
\textbf{CQ} & \textbf{Pattern}\\
\midrule  
\multirow{3}{*}{S1} & CQ1 & When is the level of a chemical substance recorded in a water body? & P1 \\
& CQ2 & What is a parameter that represents the quality of water bodies? & P2 \\
& CQ3 & Who records the amount of microbiological substances in surface waters in time? & P3 \\
\midrule  
\multirow{3}{*}{S2} & CQ4 & What are the contaminated sites in a geographical area recorded in time? & P1 \\
& CQ5 & When is the rate of hospitalisation related to a disease registered? & P2 \\
& CQ6 & Who monitors the hospitalisations for a disease in geographical area? & P3 \\
\bottomrule
\end{tabular}
\end{table}

We recruited 7 participants for the experiment with all of them being experts in ontology design and patterns. Each participant was asked to perform C1, i.e. ontology design without FrODO, with one of the two CQ sets and C2, i.e. ontology design with FrODO, with the other CQ set. The assignment of the CQ sets to the participants with respect to the two ontology design conditions was performed in order to have the same CQs in S1 and S2 modelled alternatively with FrODO or without FrODO by a a balanced number of participants. Hence, the participants were divided into two groups, i.e. (i) {\em Group 1} tackling the CQs in the set S1 under the experiment condition C2 and the CQs in the set S2 under the experiment condition C1 and (ii) {\em Group 2} tackling the CQs in the set S1 under the experiment condition C1 and the CQs in the set S2 under the experiment condition C2. Additionally, we asked 3 participants to carry out the condition C2 as first option and then the condition C1 as second option. The remaining 4 participants were asked to carry out C1 first and then C2. The rationale of this choice was to mitigate possible biases introduced in the experiment by the order that the two conditions were executed by the participants. In fact, a participant, while addressing the experiment condition requested to be the first in the sequence, might acquire, for example, a deeper knowledge of the domain or she might benefit from the re-use of one or more ontology design patterns identified and used in the condition tackled as first option. This might make the condition tackled as second option easier to be solved, thus affecting the veracity of the results. The participants were supervised by an evaluator, who was in charge of (i) introducing FrODO with a brief demonstration, (ii) providing a detailed explanation of the experiment to the participants, (iii) supporting the participants during the experiments, (iv) recording the times taken by the participants for solving conditions C1 and C2. At the end of the experiment each participant was supposed to produce two ontologies as output. That is, one ontology addressing the CQs associated with condition C1 and another ontology addressing the CQs associated with condition C2. We remark that the CQs associated with C1 or C2 were either those available for set S1 or set S2.

\begin{table}[t!]
\begin{center}
	\caption{Structural metrics used for assessing the ontologies produced by the participants to the experiment for both condition C1 and C2.}
	\label{tab:metrics}
	\begin{tabular}{p{3.8cm}p{7.5cm}}
	\toprule
	{ \bf Metric } & { \bf Description }\\
	\midrule
	\# of annotation assertions & The total number of annotations in the ontology. Values are on ordinal scale.\\
	\# of axioms & The total number of axioms defined for classes, properties, datatype definitions, assertions and annotations. Values are on ordinal scale.\\
	\# of classes & The total number of classes defined in the ontology network. Values are on ordinal scale.\\
	\# of datatype properties & The total number of datatype properties defined in the ontology network. Values are on ordinal scale.\\
	\# of inverse object properties & The total number of object properties having a \texttt{owl:inverseOf} axiom for representing their inverse properties. Values are on ordinal scale.\\
	\# of logical axioms & The axioms which affect the logical meaning of the ontology. Values are on ordinal scale.\\
	\# of object properties & The total number of object properties defined in the ontology. Values are on ordinal scale.\\
	\# of object property domain axioms & The total number of axioms specifying the domain of an object property. Values are on ordinal scale.\\
	\# of object property range axioms & The total number of axioms specifying the range of an object property. Values are on ordinal scale.\\
	\# of SubClassOf axioms & The total number of \texttt{rdfs:subClassOf} axioms defined in the ontology.\\
	Axiom/class ratio & The ratio between axioms and classes computed as the average amount of axioms per class. Values are computed as $\frac{\mbox{\# of axioms}}{\mbox{\# of classes}}$.\\
	Class/property ratio & The ratio between the number of classes and the number of properties. Values are computed as $\frac{\mbox{\# of classes}}{\mbox{\# of properties}}$.\\
	Inverse relations ratio & The ratio between the number of inverse relations and all the relations defined in the ontology. Values are on a scale ranging fron 0 to 1 and are computed as $\frac{\mbox{\# of inv. object properties} + \mbox{\# of inv. funct. datatype properties}}{\mbox{\# of object properties} + \mbox{\# of datatype properties}}$.\\
	Inheritance Richness & The average number of subclasses per class computed as proposed by~\cite{Tartir2010}. Inheritance Richness is expressed on ordinal scale.\\
	Relationship Richness & The ratio between non-inheritance relations and the total number of relations defined in the ontology as proposed by~\cite{Tartir2010}. Inheritance relations are \texttt{rdfs:subClassOf} axioms. Values are on a scale ranging from 0 (i.e. the ontology contains inheritance relationships only) to 1 (i.e. the ontology contains non-inheritance relationships only).\\
	\bottomrule
	\end{tabular}
	\end{center}
\end{table}

At the end of the experiment we asked the participants to rate ten statements using a five-point Likert scale (from 1: Strongly Disagree to 5: Strongly Agree). The ten statements are those of the System Usability Scale (SUS)~\cite{Brooke1996}. The SUS is a well-known metric used for evaluating the usability of a system. It has the advantage of being technology-independent, and reliable even with a very small sample size~\cite{Sauro2011} as in our case. It also provides a two-factors orthogonal structure, which can be used to score the scale on independent Usability and Learnability dimensions~\cite{Sauro2011}. The adoption of SUS in our experiment was not meant for assessing neither the usability of FrODO per se nor its integration with Prot\'{e}g\'{e}. On the contrary, it was meant for investigating the effectiveness of FrODO and its implemented methodology in supporting ontologists in ontology design tasks. %Table~\ref{tab:sus_open_questions} reports the four open questions aimed at collecting feedback (i.e. pros and cons) from the participants about the quality of their experience in ontology design with the support of FrODO.

%\begin{table}[!ht]\centering
%\caption{Open questions for recording participants' feedback.}
%\label{tab:sus_open_questions}
%		\begin{tabular}{ll}
%		\toprule                      
%		{\bf Nr. } & {\bf Question } \\
%		\midrule
%		 1 & How effectively did the tool support you in modelling CQs into ontologies? \\ 
%		 2 & What were the most useful features of the tool that helped you to perform your tasks? \\
%		 3 & What were the main weaknesses that the tool exhibited in supporting your tasks? \\
%		 4 & 	Would you suggest any additional features that would have helped you to accomplish your tasks? \\
%		 \bottomrule
%  	     \end{tabular}
%\end{table}

The ontologies produced by participants were evaluated with respect to the logical and structural dimensions as identified in the ontology evaluation framework formalised by~\cite{Gangemi2006}. The logical dimension was assessed by detecting the lack of inconsistencies by means of a DL reasoner. The DL reasoner we opted for is HermiT\footnote{\url{http://www.hermit- reasoner.com/}: last visited on May 2022.}. The structural dimension was assessed  with different metrics that have been defined and used in literature~\cite{Gangemi2006,Tartir2010,Lantow2016,Carriero2021}. The structural metrics we used are reported in Table~\ref{tab:metrics}. We do not assess the functional dimension, which is the ability of an ontology to address requirements and cover the domain~\cite{Gangemi2006}. This is because all the ontologies result from expert ontology engineers that were asked to model ontologies able to address the proposed CQs that identify the target ontological requirements. Hence, we assume that all resulting ontologies are qualitative from the functional perspective.
\subsection{Results}
\begin{figure}[!ht]
\begin{subfigure}[t]{1\textwidth}
  \centering
  \includegraphics[width=1\textwidth]{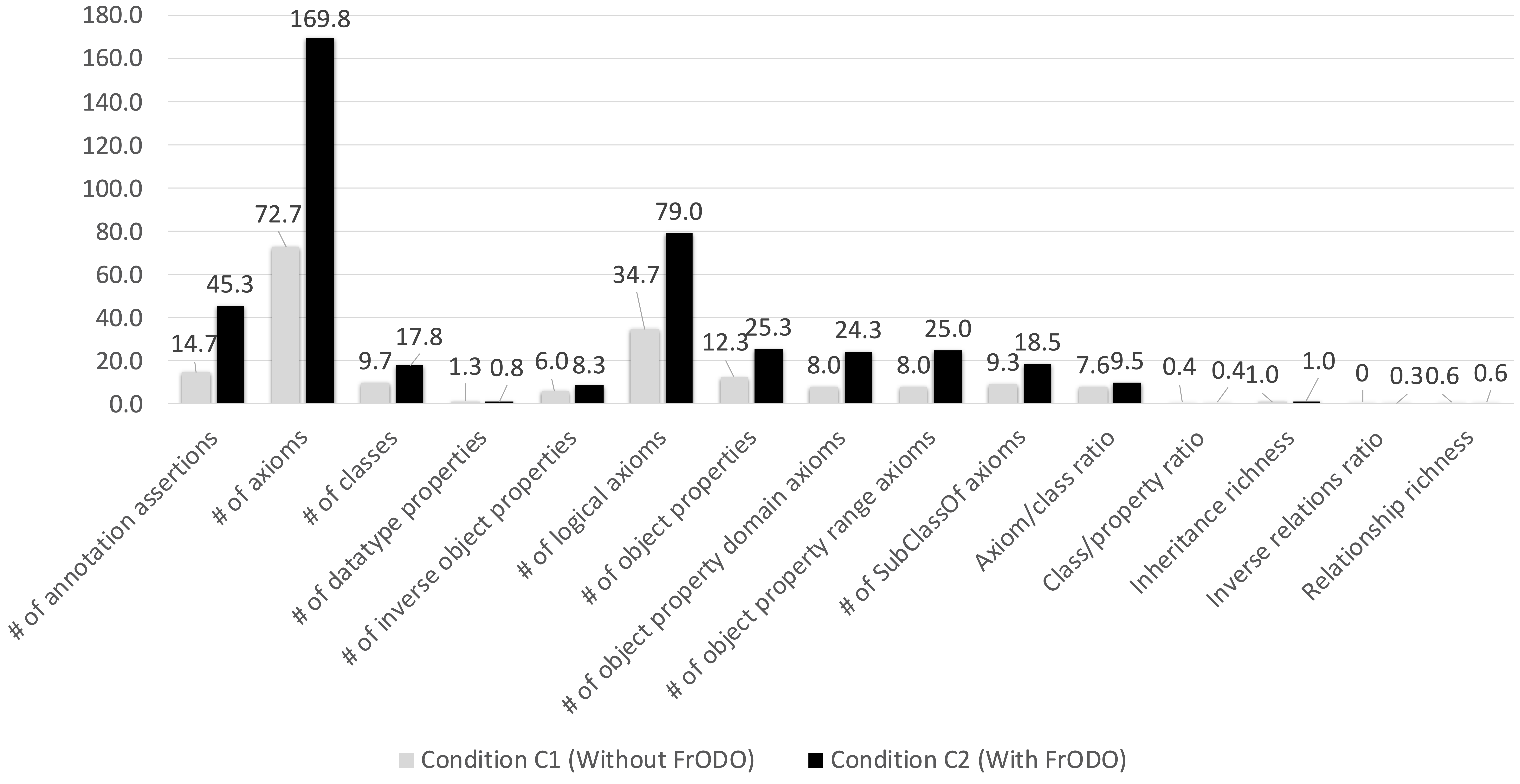}
  \caption{Group 1}
  \label{fig:results-group1}
\end{subfigure}
\begin{subfigure}[t]{1\textwidth}
  \centering
  \includegraphics[width=1\textwidth]{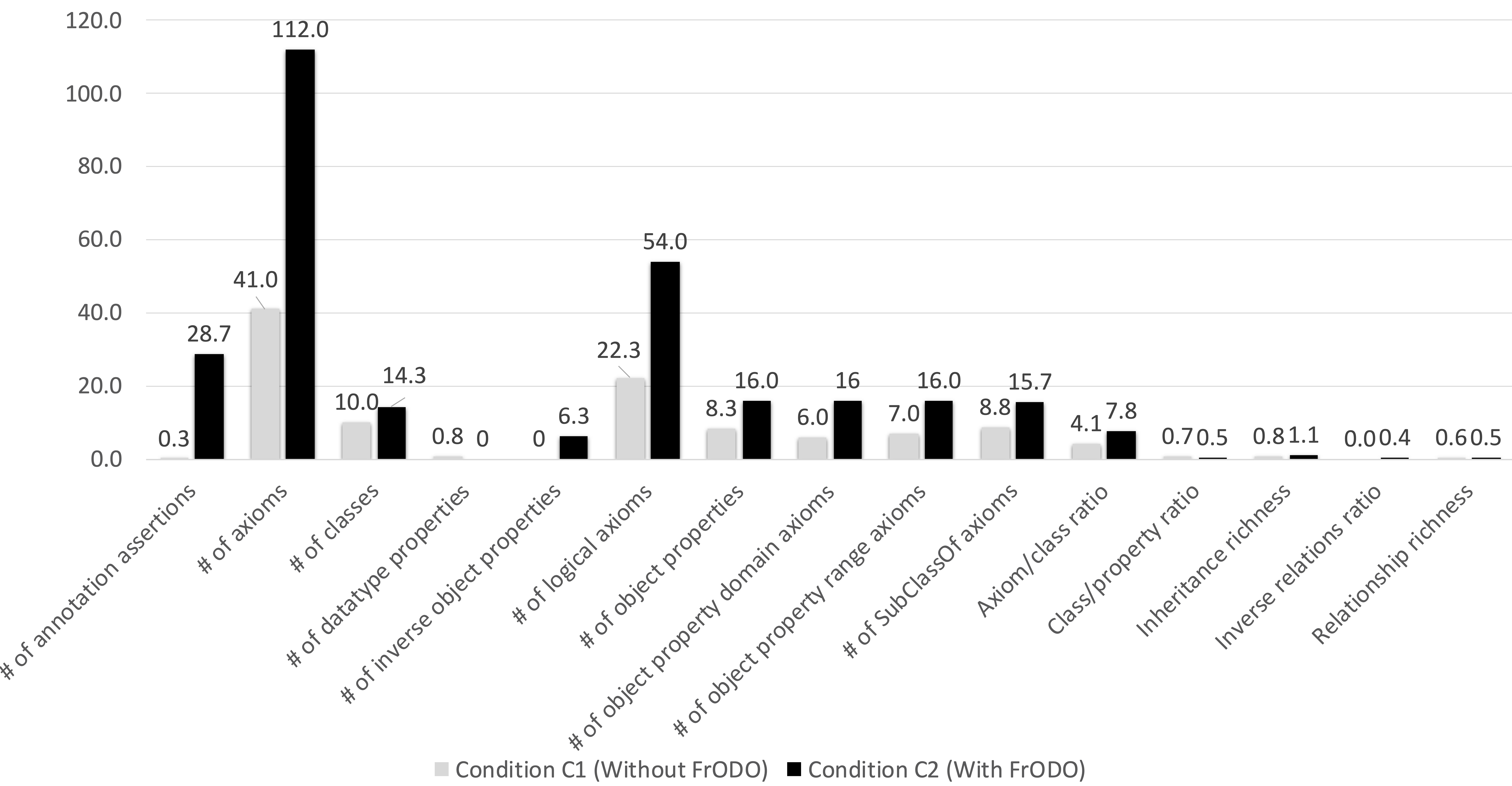}
  \caption{Group 2}
  \label{fig:results-group2}
\end{subfigure}
\caption{Results computed for all structural metrics reported in Table~\ref{tab:metrics}}
\label{fig:results}
\end{figure}
All ontologies designed by participants for conditions C1 and C2 are logically consistent. Instead, Figures~\ref{fig:results-group1} and~\ref{fig:results-group2} show the results computed for the metrics reported in Table~\ref{tab:metrics} for participant groups 1 and 2, respectively. The values reported are averaged among those obtained for each participant to the experiment with regards to the experiment condition. We used OntoMetrics\footnote{\url{https://ontometrics.informatik.uni-rostock.de/ontologymetrics/} last visited on May 2022.}~\cite{Lantow2016} as a tool for computing such metrics automatically. Most metrics we took into account perform better in the experiment condition that includes FrODO (i.e. condition C2) than in the other does not (i.e. condition C2). This observation is valid for both participant groups 1 and 2.
%For all metrics taken into account the experiment with the support of FrODO allows our participants to design richer ontologies in terms of (i) axioms, (ii) annotations, (iii) classes, (iv) taxonomic relations (i.e. \texttt{rdfs:subClassOf} relations), (v) object properties, and (vi) inverse object properties. This richness is confirmed by analysisng the results for the class/property ratio, inheritance richness, inverse relation ratio, and relationship richness. In fact, for all these metrics we recorded higher values from the experiment condition involving FrODO (i.e. condition C2) into the development process for both participant group 1 and 2.

\begin{figure}[!th]
\begin{subfigure}[t]{0.49\textwidth}
  \centering
  \includegraphics[width=1\textwidth]{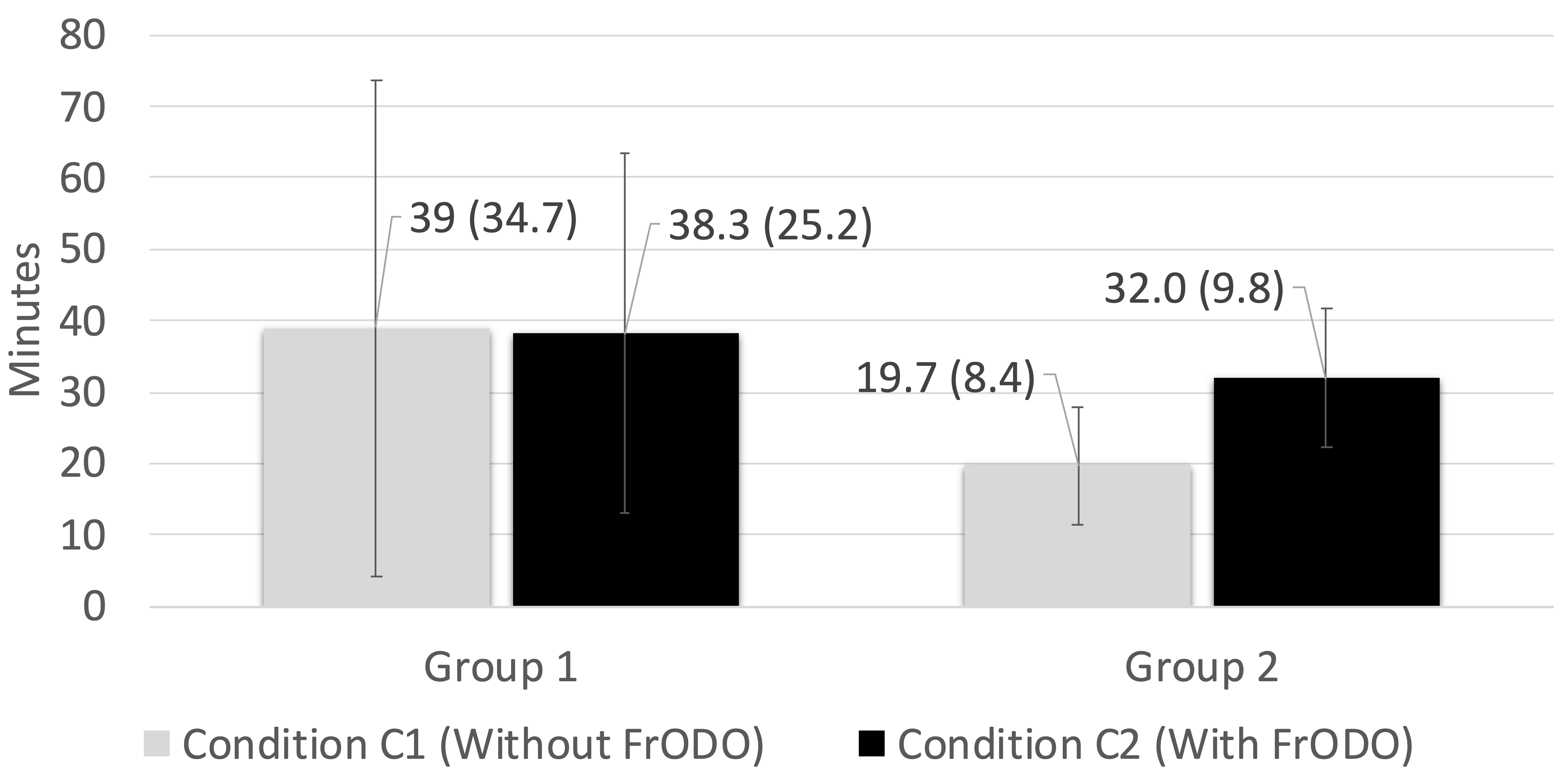}
  \caption{All participants.}
  \label{fig:times-global}
\end{subfigure}
\begin{subfigure}[t]{0.49\textwidth}
  \centering
  \includegraphics[width=1\textwidth]{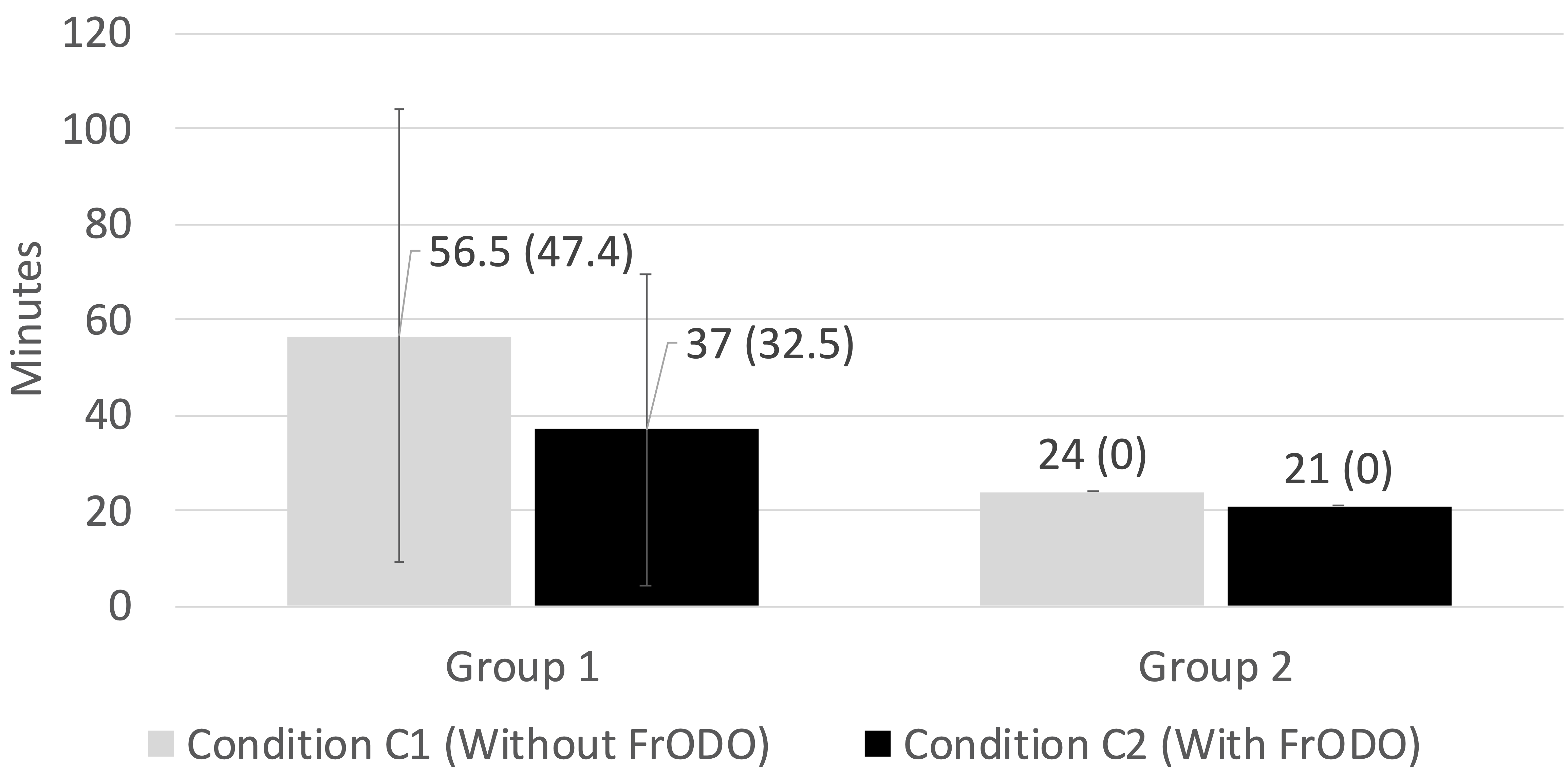}
  \caption{Participants that tackled Condition 2 after Condition 1.}
  \label{fig:times-frodo-second}
\end{subfigure}
\caption{Times required on average for completing the tasks associated with condition 1 and 2, respectively. The times are expressed in minutes.}
\label{fig:results}
\end{figure}
Figure~\ref{fig:times-global} reports the times taken on average by participants to complete the tasks associated with the conditions C1 and C2 regardless to the execution order. Standard deviation values are reported among brackets for each series. On the contrary, Figure~\ref{fig:times-frodo-second} reports the same times by taking into account the specific order in which the participants tackled C1 first and then C2. In the first case (cf. Figure~\ref{fig:times-global}) the times recorded are comparable, on average, for the participants of group 1, i.e. 39 minutes for C1 Vs 38.3 for C2. Nevertheless, for the participants of group 2 we recorded longer times when the ontology design process was supported by FrODO, i.e. 32 minutes recorded for condition C2 Vs. 19.7 minutes recorded for condition C1. However, if we limit our analysis to the case having the condition C2 executed after the condition C1 (cf. Figure~\ref{fig:times-frodo-second}), then we observe longer times when the ontology design process was not supported by FrODO. That is, for group 1, we recorded on average 37 minutes taken for C2 Vs. 56.5 taken for C1 and, for group 2, we recorded on average 21 minutes taken for C2 Vs. 24 taken for C1.

\begin{figure}[!ht]
\centering
\includegraphics[width=0.8\textwidth]{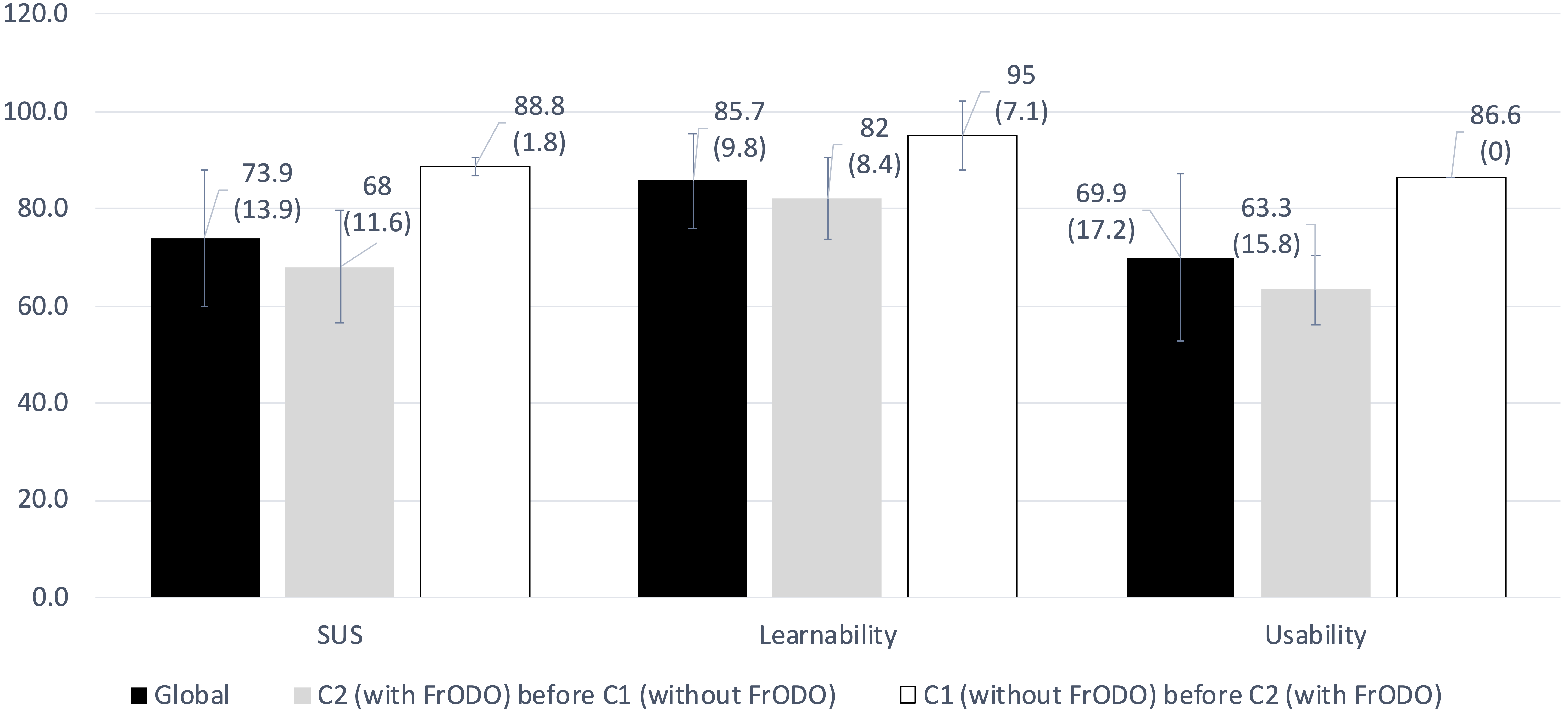}
\caption{SUS, learnability and usability scores.}
\label{fig:sus}
\end{figure}

Figure~\ref{fig:sus} reports the SUS scores in terms of the overall SUS score and its two orthogonal indicators, i.e. Learnability and Usability. Standard deviation values are reported among brackets. We report separates perspectives over the results obtained with the SUS along with the global scores by distinguishing between two cases: (a) the condition C2 (with FrODO) was executed before C1 (without FrODO) and (b) vice versa, the condition C1 (without FrODO) was executed before C2 (with FrODO). For the global perspectives the sores are 73.9, 85.87, and 69.9 for SUS, Learnability, and Usability, respectively. For perspective (a) the sores are 68, 82, and 63.3 for SUS, Learnability, and Usability, respectively. Finally, for perspective (b) the sores are 88.8, 95, and 86.6 for SUS, Learnability, and Usability, respectively.
All the ontologies generated by the participants for both conditions C1 and C2 are available on GitHub\footnote{\url{https://github.com/anuzzolese/frodo/tree/main/evaluation/ontologies}}. Simirlarly, the CSV data containing the metrics computed with OntoMetrics, times, and SUS results are published on Zenodo~\footnote{\url{https://doi.org/10.5281/zenodo.6574273}}.

\subsection{Discussion}
The analysis based on the structural metrics computed over the ontologies generated for cases C1 and C2 suggests that the ontologies produced with the support of FrODO (i.e. case C2) are richer in terms of (i) axioms, (ii) annotations, (iii) classes, (iv) taxonomic relations (i.e. \texttt{rdfs:subClassOf} relations), (v) object properties, and (vi) inverse object properties. This richness is confirmed by analysing the results for the class/property ratio, inheritance richness, inverse relation ratio, and relationship richness. In fact, for all these metrics we recorded higher values from the experiment condition involving FrODO (i.e. condition C2) into the development process for both participant groups 1 and 2. This addresses {\em RQ1}.
The SUS-based analysis shows that an ontology design methodology based on FrODO can be considered usable regardless to the specific order in which conditions C1 and C2 were set in the experiment. In fact, based on empirical studies [46], a SUS score of 68 represents the average usability value even in case of a small number of participants~\cite{Sauro2011}. Additionally, we observed, on average, that the adoption of FrODO in an ontology design workflow is not time-consuming. This is satisfactory and addresses {\em RQ2}.

\section{Conclusions and future work}
\label{sec:concl}
In this work we present the Frame-based Ontology Design Outlet (FrODO), which is a method and tool able to draft ontologies from competency questions (CQs) automatically. First, we provide a background about ontology design methodologies and automatic solutions for generating ontologies from text. Then, we describe the method implemented by FrODO that builds on top of FRED for leveraging frame semantics to gather domain knowledge and formalise it into OWL ontology drafts. Those drafts can be used by ontology engineers to make agile ontology design methodologies (e.g. XD, SAMOD, etc.) smoother. The effectiveness of FrODO was assessed by means of a user-based study that involved 7 participants all of them being expert ontology engineers. The aim of the a study was twofold, that is, evaluating: (i) the quality of the ontology drafts produced by FrODO from the logical and structural perspectives; and (ii) the usability of FrODO when used for an ontology design task. The experiment shows that FrODO produces richer ontologies if compared to the ontologies designed without the support of FrODO from the same set of CQs. The System Usability Scale (SUS) shows excellent usability scores for all the perspective we analysed. Future works include the evaluation of the ontologies produced with the support of FrODO from the functional perspective. Then, we aim at designing and developing a plug-in for Prot\'{e}g\'{e} able to embed FrODO inside the popular ontology design framework, thus strengthening the cohesion between the two systems. Additionally, we plan to extend FrODO in order to provide better support to the generation of datatype properties, inverse functional datatype properties, and disjoint axioms, which are overlooked in the current version of the tool. Among the others the implementation of disjoint axioms will benefit from the alignment with foundational ontologies, such as DOLCE+DnS Ultralite, which is already provided by FRED.

\section*{Acknowledgements}
This work has been supported by the Water Health Open knoWledge (WHOW) project co-financed by the Connecting European Facility programme of the European Union under grant agreement INEA/CEF/ICT/A2019/206322. 

\bibliographystyle{vancouver}
\bibliography{references}

\begin{thebibliography}{10}

\bibitem{Gruninger1995}
Gr{\"u}ninger M, Fox MS.
\newblock The role of competency questions in enterprise engineering.
\newblock In: Benchmarking—Theory and practice. Springer; 1995. p. 22-31.

\bibitem{Presutti2009}
Presutti V, Daga E, Gangemi A, Blomqvist E.
\newblock eXtreme Design with Content Ontology Design Patterns.
\newblock In: Blomqvist E, Sandkuhl K, Scharffe F, Svátek V, editors. Proc. of
  WOP 2009. vol. 516 of CEUR Workshop Proceedings. CEUR-WS.org; 2009. .

\bibitem{SAMOD2016}
Peroni S.
\newblock A Simplified Agile Methodology for Ontology Development.
\newblock In: Dragoni M, Poveda-Villalón M, Jiménez-Ruiz E, editors. Proc of
  OWLED 2016. vol. 10161 of Lecture Notes in Computer Science. Springer; 2016.
  p. 55-69.
\newblock DOI: 10.6084/M9.FIGSHARE.3189769.V2.

\bibitem{Gangemi2017}
Gangemi A, Presutti V, Recupero DR, Nuzzolese AG, Draicchio F, Mongiovì M.
\newblock Semantic Web Machine Reading with FRED.
\newblock Semantic Web. 2017;8(6):873-93.
\newblock DOI: 10.3233/SW-160240.

\bibitem{Etzioni2007}
Etzioni O, Banko M, Cafarella MJ.
\newblock Machine Reading.
\newblock In: AAAI Spring Symposium: Machine Reading. AAAI; 2007. p. 1-55.
\newblock DOI: 10.5555/1597348.1597430.

\bibitem{Fillmore2006}
Fillmore CJ, et~al.
\newblock Frame semantics.
\newblock Cognitive linguistics: Basic readings. 2006;34:373-400.

\bibitem{Gangemi2009}
Gangemi A, Presutti V.
\newblock Ontology design patterns.
\newblock In: Handbook on ontologies. Springer; 2009. p. 221-43.

\bibitem{GrunFox1995}
Gr\"{u}ninger M, Fox M.
\newblock Methodology for the Design and Evaluation of Ontologies.
\newblock In: IJCAI'95, Workshop on Basic Ontological Issues in Knowledge
  Sharing; 1995. .

\bibitem{Sure2004}
Sure Y, Staab S, Studer R.
\newblock In: Staab S, Studer R, editors. On-To-Knowledge Methodology (OTKM).
  Berlin, Heidelberg: Springer Berlin Heidelberg; 2004. p. 117-32.
\newblock DOI: 10.1007/978-3-540-24750-0\_6.

\bibitem{Keet2016}
Keet CM, Lawrynowicz A.
\newblock Test-Driven Development of Ontologies.
\newblock In: Sack H, Blomqvist E, d'Aquin M, Ghidini C, Ponzetto SP, Lange C,
  editors. ESWC. vol. 9678 of Lecture Notes in Computer Science. Springer;
  2016. p. 642-57.
\newblock DOI: 10.1007/978-3-319-34129-3\_39.

\bibitem{Wizniewski2019}
Wiśniewski D, Potoniec J, Ławrynowicz A, Keet C.
\newblock Analysis of Ontology Competency Questions and their formalisations in
  SPARQL-OWL.
\newblock Journal of Web Semantics. 2019 11;59.
\newblock DOI: 10.1016/j.websem.2019.100534.

\bibitem{Carriero2019}
Carriero VA, Mariani F, Nuzzolese AG, Pasqual V, Presutti V.
\newblock Agile Knowledge Graph Testing with TESTaLOD.
\newblock In: ISWC Satellites; 2019. p. 221-4.

\bibitem{Wisniewski2018}
Wisniewski D.
\newblock Automatic translation of competency questions into sparql-owl
  queries.
\newblock In: Companion Proceedings of the The Web Conference 2018; 2018. p.
  855-9.
\newblock DOI: 10.1145/3184558.3186575.

\bibitem{Cimiano2006}
Cimiano P.
\newblock Ontology learning and population from text: algorithms, evaluation
  and applications. vol.~27.
\newblock Springer Science \& Business Media; 2006.

\bibitem{Al2020}
Al-Aswadi FN, Chan HY, Gan KH.
\newblock Automatic ontology construction from text: a review from shallow to
  deep learning trend.
\newblock Artificial Intelligence Review. 2020;53(6):3901-28.
\newblock DOI: 10.1007/s10462-019-09782-9.

\bibitem{Cimiano2005}
Cimiano P, V{\"o}lker J.
\newblock A framework for ontology learning and data-driven change discovery.
\newblock In: Proceedings of the 10th International Conference on Applications
  of Natural Language to Information Systems (NLDB). Springer; 2005. p. 227-38.
\newblock DOI: 10.1007/11428817\_21.

\bibitem{Witte2010}
Witte R, Khamis N, Rilling J.
\newblock Flexible ontology population from text: The owlexporter.
\newblock In: Proceedings of LREC 2010; 2010. .

\bibitem{Tanev2006}
Tanev H, Magnini B.
\newblock Weakly supervised approaches for ontology population.
\newblock In: Proceedings of 11th Conference of the European Chapter of the
  Association for Computational Linguistics; 2006. p. 17-24.

\bibitem{Hearst1992}
Hearst M.
\newblock Automatic acquisition of hyponyms from large text corpora in proc.
\newblock In: 14th International Conference Computational Linguistics, Nantes
  France; 1992. .

\bibitem{Voelker2008}
V\"{o}lker J, Rudolph S.
\newblock Lexico-Logical Acquisition of OWL DL Axioms - An Integrated Approach
  to Ontology Refinement.
\newblock In: Proceedings of ICFCA 2008. vol. 4933 of Lecture Notes in
  Artificial Intelligence; 2008. .

\bibitem{Falco2014}
Falco R, Gangemi A, Peroni S, Shotton DM, Vitali F.
\newblock Modelling OWL Ontologies with Graffoo.
\newblock In: Presutti V, Blomqvist E, Troncy R, Sack H, Papadakis I, Tordai A,
  editors. ESWC (Satellite Events). vol. 8798 of Lecture Notes in Computer
  Science. Springer; 2014. p. 320-5.
\newblock 10.1007/978-3-319-11955-7\_42.

\bibitem{Ren2014}
Ren Y, Parvizi A, Mellish C, Pan JZ, Deemter Kv, Stevens R.
\newblock Towards competency question-driven ontology authoring.
\newblock In: European Semantic Web Conference. Springer; 2014. p. 752-67.

\bibitem{Tartir2010}
Tartir S, Arpinar IB, Sheth AP.
\newblock Ontological evaluation and validation.
\newblock In: Theory and applications of ontology: Computer applications.
  Springer; 2010. p. 115-30.
\newblock DOI: 10.1007/978-90-481-8847-5\_5.

\bibitem{Brooke1996}
Brooke J.
\newblock {SUS - A quick and dirty usability scale}.
\newblock Usability evaluation in industry. 1996;189(194):4-7.
\newblock DOI: 10.1201/9781498710411-35.

\bibitem{Sauro2011}
Sauro J.
\newblock A practical guide to the system usability scale: Background,
  benchmarks \& best practices.
\newblock Measuring Usability LCC; 2011.

\bibitem{Gangemi2006}
Gangemi A, Catenacci C, Ciaramita M, Lehmann J.
\newblock Modelling ontology evaluation and validation.
\newblock In: European Semantic Web Conference. Springer; 2006. p. 140-54.
\newblock DOI: 10.1007/11762256\_13.

\bibitem{Lantow2016}
Lantow B.
\newblock OntoMetrics: Putting Metrics into Use for Ontology Evaluation.
\newblock In: KEOD; 2016. p. 186-91.
\newblock DOI: 10.5220/0006084601860191.

\bibitem{Carriero2021}
Carriero VA, Gangemi A, Mancinelli ML, Nuzzolese AG, Presutti V, Veninata C.
\newblock Pattern-based design applied to cultural heritage knowledge graphs.
\newblock Semantic Web. 2021;12(2):313-57.
\newblock DOI: 10.3233/SW-200422.

\end{thebibliography}
\end{document}